%% file: main.tex
\documentclass[10pt,twocolumn,letterpaper]{article}

\usepackage{cvpr}

\usepackage{times}
\usepackage{relsize}
\usepackage[T1]{fontenc} 
\usepackage[latin1]{inputenc} 
\usepackage[english]{babel}

\usepackage{epsfig}
\usepackage{graphicx}
\usepackage{wrapfig}
\usepackage[belowskip=0pt,aboveskip=0pt,font=small]{caption}
\usepackage{subcaption}
\usepackage{amsmath, amsthm, amssymb}
\usepackage{textcomp}
\usepackage{upgreek}
\usepackage{bm}
\usepackage{cases}
\usepackage{mathtools}

\usepackage{cite}
\usepackage[pagebackref=true,breaklinks=true,letterpaper=true,colorlinks,bookmarks=false,citecolor=red]{hyperref}

\usepackage{multirow}
\usepackage{rotating}
\usepackage{booktabs}

\usepackage{enumitem}
\usepackage[olditem,oldenum]{paralist}

\usepackage{alltt}
\usepackage{listings}

\usepackage{mysymbols}

\usepackage{url}
\usepackage{xspace}
\usepackage{comment}
\usepackage{color}
\usepackage{afterpage}
\usepackage{pdfpages}
\usepackage{framed}
\usepackage{fancybox}

\usepackage{tabularx}
\usepackage{longtable}
\usepackage{tabu}
\usepackage{dblfloatfix}
\usepackage{overpic}
\usepackage{standalone}
\usepackage{titling}

\input definitions.tex

\cvprfinalcopy 
\def\cvprPaperID{240} 

\begin{document}

\title{Combining the Best of Graphical Models and ConvNets\\ for Semantic Segmentation}

\author{Michael Cogswell \\
Virginia Tech\\
{\tt\small cogswell@vt.edu}
\and
Xiao Lin \\
Virginia Tech\\
{\tt\small linxiao@vt.edu}
\and
Senthil Purushwalkam \\
Virginia Tech\\
{\tt\small senthil@vt.edu}
\and
Dhruv Batra \\
Virginia Tech\\
{\tt\small dbatra@vt.edu}
}

\date{}

%
%
%

\maketitle


\begin{abstract} 
We present a two-module approach to semantic segmentation that
incorporates Convolutional Networks (CNNs) and Graphical Models. 
Graphical models are used to generate a small (5-30) set of diverse 
segmentations proposals, such that this set has high recall. 
Since the number of required proposals is so low, we can
extract fairly complex features to rank them. 
Our complex feature of choice is a novel 
CNN called \segnet, which 
directly outputs a (coarse) semantic segmentation. 
Importantly, \segnet is specifically trained to optimize the corpus-level PASCAL
IOU loss function. To the best of our knowledge, this is the first CNN 
specifically designed for semantic segmentation. 
This two-module approach achieves $52.5\%$ on the PASCAL 2012
segmentation challenge.
\end{abstract}

\input{intro}

\input{related_work}

\input{approach}

\input{results}

\input{conclusion}


\nocite{jia13caffe}
{ 
\bibliographystyle{ieee}
\bibliography{bib_file}
}

\pagebreak
\setcounter{section}{0}
\onecolumn
\input{uoi_derivative/uoi_loss_gradient_innards.tex}

\end{document}

%% file: definitions.tex
\newcommand{\segnet}{$\mathtt{SegNet}$\xspace}
\newcommand{\segnets}{$\mathtt{SegNets}$\xspace}

\newcommand{\conv}{$\mathtt{conv}$\xspace}
\newcommand{\pool}{$\mathtt{pool}$\xspace}
\newcommand{\fc}{$\mathtt{fc}$\xspace}

\newcommand{\divmbest}{DivMBest\xspace}

\newcommand{\reranking}{re-ranking\xspace}
\newcommand{\Reranking}{Re-ranking\xspace}
\newcommand{\reranker}{re-ranker\xspace}

\newcommand{\rerank}{re-rank\xspace}

\newcommand{\oracle}{\texttt{oracle}\xspace}

\newcommand{\pp}{\%-points\xspace}

\newcommand{\val}{$\mathtt{val}$\xspace}
\newcommand{\trainval}{$\mathtt{trainval}$\xspace}
\newcommand{\test}{$\mathtt{test}$\xspace}

%% file: intro.tex
\section{Introduction}
\label{sec:intro}

Training deep Convolutional Neural Networks (CNNs) with large amounts of labeled 
data has produced impressive results for classification and detection of 
objects and attributes~\cite{krizhevsky2012imagenet, 
girshick2013convnet, sermanet2013overfeat, zhang2013panda}.
The natural next question to ask is -- can these deep models be generalized
beyond simple prediction spaces (as in multi-way classification) 
to complex, structured prediction spaces as in 
semantic segmentation, keypoint/pose estimation, and coarse 3D estimation?


There are two main challenges in this generalization: 
\begin{compactitem}

\item \textbf{Does vision $=$ ``lots of classification''?}
Most recent applications of CNNs to new tasks 
such as detection and segmentation have framed these tasks as 
``lots of classification'', 
either of scanning window patches~\cite{farabet2013pami, ciresan2012deep, pinheiro2014rcnn, 
grangier2009deep, schulz2012cnn} or region 
proposals \cite{girshick2013convnet,hariharan2014sds,he2014spatial}. 
While these results are encouraging,  
such formulations ignore the rich structure in the output space. 
In semantic segmentation, the goal is to label each pixel 
with an object class. Labels of nearby pixels tend to be correlated, 
and independent per-pixel predictions loose this valuable signal. 
These intuitions are also reflected in the choice of the evaluation metrics 
used by community -- for instance, mean Jaccard Index (or Intersection-over-Union (IOU)) 
used by PASCAL segmentation, as opposed to the \naive Hamming distance. 


\item \textbf{Limited training data.}
Unlike classification, which requires image-level labels, and detection, 
which requires bounding boxes, higher-level scene understanding tasks such as
semantic segmentation, or coarse 3D estimation often require
\emph{dense pixel-level} annotations that are time consuming and 
expensive to collect. Thus, such datasets are significantly smaller
in scale than classification, despite ongoing valiant efforts~\cite{coco}. 

\end{compactitem}

\input{figs/teaser/teaser.tex}

\textbf{Goal.}
At a high level, the goal of this paper is to address the above two
challenges -- to leverage improvements in CNN-based classification for
higher-level vision tasks in a manner that uses the large training corpus
available for classification without ``shoe-horning'' the task at hand into
repeated classification. 

\textbf{Overview.} 
We present a novel CNN-based approach for semantic segmentation, the task of 
labeling each pixel in an image with an object class. 
\figref{fig:teaser} illustrates our two-module approach. 
Module 1 uses a graphical model to produce 
multiple semantic segmentation proposals. 
Module 2 uses a novel CNN called \textbf{\segnet}, which 
is used to score and \rerank these proposals, resulting in the final prediction.


\textbf{Contributions.}
%
Our primary technical contribution is \segnet, a novel CNN that
directly outputs a (coarse) semantic segmentation. Importantly, \segnet is
task-aware, and \emph{specifically trained to optimize the corpus-level PASCAL
IOU loss function}.
To the best of our knowledge, this is the first CNN 
specifically designed for semantic segmentation. 
%
While our experiments focus on this one specific application (semantic segmentation), 
at a high-level, our approach presents a general recipe for combining 
the strengths of graphical models (modeling dependencies) and deep learning (learning rich features) 
for a range of applications. The recipe is simple -- 
use graphical models to generate a small set of proposals and CNNs to score them. 
%
%
%
Formulating the problem this way has a number of advantages:
\begin{compactitem}

\item \textbf{Wider receptive field without loss in resolution:}
As CNNs get deeper, each output pixel gets to see a larger patch of input and reason 
about more context.
Unfortunately, the output also gets coarser due to the pooling layers.
Thus, practitioners are left with a dilemma -- either build shallow networks that 
have limited performance or deeper richer networks that loose localization information. 
Our 2-module approach does not face this problem; 
\segnet gets to look at not just a patch or a segment, but the 
entire image to make its predictions. The loss in resolution is acceptable 
because the \segnet prediction simply needs to \rerank holistic proposals, which are full resolution. 

\item 
\textbf{Leveraging classification corpus while learning output structure:} 
The first few layers of \segnet are warm-started with 
with Krizhevsky~\etal's classification network (AlexNet) trained on ImageNet \cite{krizhevsky2012imagenet}.
These weights have learned the expected Gabor-like filters, and are good low-level features 
for natural images. 
We make the last few layers task-aware by optimizing corpus-level structured loss on PASCAL.  

\item \textbf{Graphical models encode knowledge about output structure:}
\Reranking proposals produced by graphical models allows us to 
reason about segmentation structure in a \emph{second} way -- 
through the large body of work tying graphical models and structured prediction.

\end{compactitem}

%% file: figs/teaser/teaser.tex
\begin{figure*}

\includegraphics[width=\textwidth]{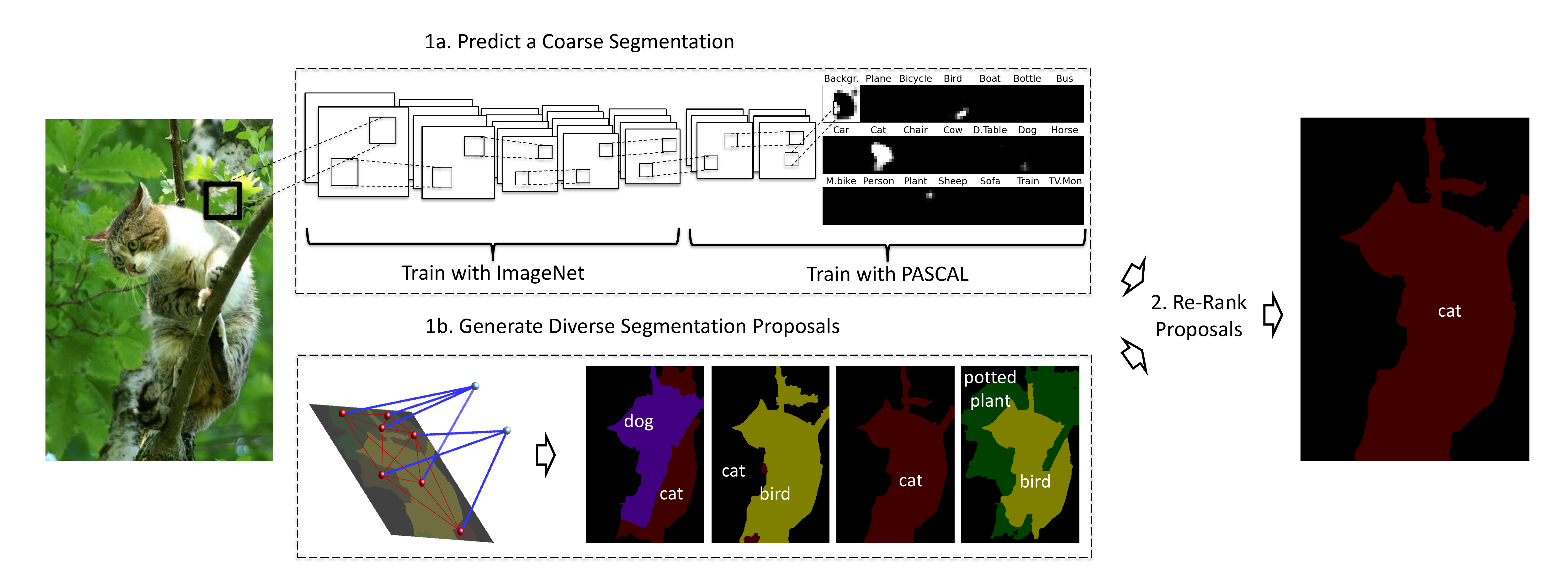}
    \caption{In 1a We predict a coarse image labeling directly from our
Convolutional Neural Network called SegNet while, in parallel, 1b uses
a CRF to predict diverse segmentation proposals. Part 2 combines these modules
by learning rank the proposals using SegNet.}
\label{fig:teaser}
\end{figure*}

%% file: related_work.tex
\section{Related Work}


As \figref{fig:teaser} suggests, 
our work relates to two themes -- deep learning and proposal-based vision pipelines.

\textbf{Convolutional Neural Networks.}
Image classification and object detection are formulated as patch 
classification problems, where the patch is either the whole image or comes from
a set of boxes sampled across scale and aspect ratio.
Segmentation is a natural extension of this view, and a number of recent approaches have
classified uniform patches sampled in a grid 
\cite{farabet2013pami, ciresan2012deep, pinheiro2014rcnn, grangier2009deep,
schulz2012cnn}.

The size of the patch in consideration (or ``receptive field'') determines the 
amount of context available -- 
\cite{farabet2013pami} and \cite{schulz2012cnn} use 
multi-scale CNNs to increase the receptive field while limiting increase in 
model complexity; \cite{schulz2012cnn} simply makes their convolution filters
larger. Most notably, \cite{pinheiro2014rcnn} uses a
recurrent CNN 
to gain depth and a larger receptive field, 
while limiting the parameters that need be learned. 
Our network is deeper, so our receptive field is naturally large.

Interestingly, a number of these approaches find that the structure 
in natural images isn't well respected by their CNN predictions. 
Thus, the CNN predictions are post-processed using graphical models 
\cite{farabet2013pami} to insert structural knowledge back into the pipeline. 
To contrast, our proposal \reranking step can be thought of as a sophisticated form of 
post-processing.

\nocite{mussi2007object}
\nocite{gattaunrolling}
\nocite{sermanet2009multirange}

\textbf{Graphical Models and Proposals.}
Modern approaches for object detection and semantic segmentation increasingly rely 
on category independent bounding-box and segment proposals
\cite{carreira_cvpr10, uijlings2013selective, arbelaez2014mcg}.
In both cases, the search space (\#boxes, \#segments) is overwhelmingly large, 
and the goal is to reduce the search space to enable expensive processing, 
without throwing out good solutions. 
\cite{girshick2013convnet} and \cite{hariharan2014sds} achieved state of the art performance on 
detection and segmentation respectively 
by classifying bounding-box and region proposals using a CNN. 
%

Most proposal methods need to produce on the order of 
200-5000 proposals to get sufficiently high recall. 
Our approach may be viewed as an instantiation of the same philosophy, 
only operating a step ``downstream''. 
Specifically, we produce entire image labelings, not category-independent box/segment proposals. 
Interestingly, this allows us to use significantly fewer proposals -- 
on the order of 10-30 per image. 
We are motivated by the 
observation made in recent work~\cite{yadollahpour2013rerank} 
-- \emph{a set of just 10 image labelings has the potential to improve PASCAL 
segmentation by $15$\pp ($33\%$ relative gain)}. 
Using fewer proposals allows even more complex scoring of those proposals 
by sophisticated secondary modules, as we do in this work. 


In a manner similar to us, 
the most successful detection and segmentation methods, 
do \emph{not} use their CNNs for localization; rather the CNNs are used to score 
proposals~\cite{girshick2013convnet, hariharan2014sds}.
On the other hand, CNNs generate dense features in
~\cite{farabet2013pami,sermanet2013overfeat}, but are outperformed by proposal-based
methods.

%% file: approach.tex
\section{Approach}
\label{sec:approach}

We begin by describing the \segnet module in our approach, and 
then explain how it is used to score semantic segmentation proposals.

\subsection{\segnet: Predicting Coarse Segmentations}
\textbf{Architecture.}
As shown in \figref{fig:teaser}, our architecture contains 8 convolutional
layers and each is fed through rectified linear non-linearities except the last,
which is fed through a pixel-wise $C$-way softmax to label an image with
$C$ classes.
There are no fully connected layers.

\textbf{Comparison with a classification net.} 
The first 5 layers in \segnet are convolutional layers (\conv), 
with 96, 256, 384, 384, then 256 filters,
Max pooling (\pool) and local response normalization follow the first
two layers, similar to AlexNet.
\footnote{More specifically, CaffeNet \cite{jia13caffe}, which is AlexNet, 
except Local Response Normalization and Max
Pooling layers are swaped. Differences are summarized in 
\url{https://github.com/BVLC/caffe/issues/296}.}
The 5th \conv layer produces 256 feature maps of size $13 \times 13$, but
we do not pool after this layer, 
and 
this is where the architectures diverge. In standard classification CNN, 
\conv layers are typically followed by fully-connected (\fc) 
layers. 
We do not have \fc layers. 
Instead, we add two more \conv layers with 128 feature maps
each and a third \conv layer with as many feature maps as the number of classes 
(including the `background' class). 
These $C$ final feature maps can be thought of as `semantic feature maps' since 
they give pixel-wise probabilities for each class and those are interpretable.

Crucially, we initialize the first 5 \conv layers with weights from CaffeNet  
trained on ImageNet, thus utilizing the large classification corpus. 
During training, we keep these \conv-layer weights fixed 
and only learn the weights of the newly added layers. 
We apply dropout \cite{hinton2012dropout} before each of the added 
feature maps during training. By feeding each pixel at the output
through a softmax activation function (normalized over classes)
we can output `semantic feature maps' which collectively give
a distribution over classes at each output pixel in the 13 $\times$ 13 grid.
An example which shows 21 feature maps for the 21 PASCAL classes is shown in
\figref{fig:coarse}.

Since \segnet contains no fully-connected layers,
the only weights are the filters. 
This is greatly beneficial since a majority of the parameters in standard classification nets 
lie in the fully-connected layers. 
Indeed, \segnet contains less than 10\% of CaffeNet parameters. 
Interestingly, in previous work, Zeiler and Fergus \cite{zeiler2013convnet} have observed that 
weights in convolutional filters provide more information per weight because
removing fully connected layers (containing most parameters) 
does not lead to a proportional decline in classification performance. 

Since \segnet predictions are coarse (low-resolution), 
we need to down-sample high-resolution segmentation ground truths to derive 
the annotation for training \segnet parameters. Each pixel in a down-sampled 
segmentation corresponds to a patch in the high-res version, so
we compute distributions over classes in this patch, yielding a soft segmentation 
ground-truth, similar to our predictions. 

Our baseline loss for training \segnet is the standard cross-entropy computed 
between a pixel's predicted class distribution and the ground truth's distribution. 
Notice that this loss function is ``decomposable'' over pixels -- it 
treats segmentation as independent classification problems at each pixel.

\subsection{Optimizing a Segmentation-Specific Loss}

Recall that the standard evaluation criteria used in segmentation tasks is 
Intersection-over-Union(IOU) averaged across classes. 
Although imperfect (in the sense that it does not reward boundary alignment), 
it does captures some notions of a good segmentation better than decomposable metrics such as 
Hamming. 
Unfortunately, this metric does not decompose over pixels or even images. 
In fact, it is a \emph{corpus-level} metric, and can only be computed for 
an entire dataset, not individual images. 
Fortunately, we only need a loss's gradient to train a CNN, so we can directly
optimize such a metric. The supplementary material shows our our derivation 
of IOU's gradient. 
Before going further, it's worth taking a detailed look at how
this loss behaves when optimized via gradient descent.

Consider high-resolution predictions and ground truth. 
Let $TP_k$ denote the number of true positive for class $k$ across the dataset, 
\ie the number of pixels across all images that are annotated and predicted as class $k$. 
Analogously, let $FP_k$ denote the false positives, 
$FN_k$ the false negatives, and $GT_k$ the sum of ground truth 
pixels for class $k$. 
Then, the Jaccard Index for class $k$ can be defined as:
\begin{align}
IOU_k = \frac{TP_k}{TP_k+FP_k+FN_k}, 
\end{align}
which is averaged across categories to yield the final metric: 
%
\begin{align}
IOU &= \frac{1}{K}\sum_{k=1}^K IOU_k 
\end{align}
%
It might seem intuitive to aim to maximize IOU, but in our experiments we 
have found the optimization to be easier when we minimized the Union-over-Intersection (UOI) instead. In our preliminary experiments with random initialisation, the IOU objective function always led to an all-background prediction whereas UOI minimization leads to a much better solution.
We explain why this might be the case. 

\input{figs/fig4}

\textbf{Optimizing IOU.}
First, let us use the fact that $TP_k + FN_k = GT_k$ to 
rewrite the gain function as: 
\begin{align}
IOU_k &= \frac{GT_k-FN_k}{GT_k+FP_k}
\end{align}
For the sake of building an understanding, consider the gradients of the 
$IOU_k$ gain function 
with respect to the two kinds of mistakes ($FP_k$ and $FN_k$): 
\begin{align} 
\frac{\partial (IOU_k)}{\partial (FP_k)} &= \frac{-(GT_k-FN_k)}{(GT_k+FP_k)^2}\\
\frac{\partial (IOU_k)}{\partial (FN_k)} &= \frac{-1}{GT_k+FP_k}
\end{align}
Notice that there are two things non-ideal about these gradients. 
First, as the number of mistakes ($FP_k$ or $FN_k$) increase the gradients \emph{diminish}. 
Second, as the number of mistakes reduce, the gradients \emph{increase}. 
Such a behavior hampers convergence of first-order methods. 
%
%

%
%
%

Note that we only analyzed the effect of $FP_k, FN_k$ on $IOU_k$, and but not on 
$IOU_{k'}$ for other categories $k'$. 
Each pixel can be assigned only one 
category, and thus the mistakes $\{FP_k, FN_k\}_1^K$ are not independent of each other. 
Thus, we need to also analyze the other terms $\del(IOU_{k'})/\del(FP_k)$. 
\figref{fig:sim} shows a simulation where we computed the 
behavior of IOU as a function of increasing $FP_k$ and $FN_k$. 
Our illustration of the gradients of $IOU_k$ provides an intuition for the behavior of the IOU function. 


\textbf{Optimizing UOI.}
Now we show that Union-over-Intersection (UOI) is a smoother optimization
function based on the behavior of its gradient and that it shares a 
natural relation with IOU. In a manner similar to IOU, the $UOI$ function can be written as:
\begin{align}
UOI &= \frac{1}{K}\sum_{k=1}^{K} UOI_k\\
&= \frac{1}{K}\sum_{k=1}^{K} \frac{TP_k + FP_k + FN_k}{TP_k}\\
&= \frac{1}{K}\sum_{k=1}^{K} \frac{GT_k+FP_k}{GT_k-FN_k}
\end{align}
Consider the gradient of  $UOI_k$ \wrt the number of mistakes $FP_k$ and $FN_k$:
\begin{align}
\frac{\partial UOI_k}{\partial FP_k} &= \frac{1}{GT_k - FN_k}\\
\frac{\partial UOI_k}{\partial FN_k} &= \frac{GT_k+FP_k}{(GT_k - FN_k)^2}
\end{align}

We can see that $UOI_k$ has more desirable properties compared to $IOU_k$, 
as illustrated in \figref{fig:sim}. 
When the number of mistakes are large, the gradient is large. 
As the number of mistakes decrease, the gradient decreases as well. 


\textbf{Does UOI optimise IOU?}
%
Since we have now established that the UOI function has more desirable traits,
we should understand whether the two objectives are related. Does minimization
of UOI lead to the maximization of IOU?
	
We show that IOU can be \emph{lower-bounded} by a decreasing function 
of UOI: 
\begin{equation}
    \sum_k IOU_k \ge f\left(\sum_k UOI_k\right). 
\end{equation}
Since $f(x)$ is a decreasing function in $x$, 
we can see that decreasing UOI leads to 
increasing the lower-bound on IOU. 
Importantly, we can show that the bound is \emph{tight} -- 
that the maximum possible value of $\sum_k IOU_k$ (=K) is achievable, 
although it requires that $TP_k \ne 0$ for all classes $k$. 
In our coarse segmentation setting this can always be achieved by not allowing any 
soft outputs to be 0. 

\begin{proof}
Assume $0 < IOU_i$.
Now consider the decreasing function $f(x) = \frac{1}{x}$. 
We need to show that:
\begin{enumerate}
\item $\sum_k IOU_k \ge \frac{1}{\sum_k UOI_k}$
\item There exists a value of UOI for which IOU=1
\end{enumerate}

Let $x_i = IOU_i$. Notice that $0 < x_i \le 1$. This means that: 
\begin{align}
    \sum_{k=1}^K x_k  \ge x_i 
    \quad
\Rightarrow 
\quad
   \frac{1}{x_i}  \ge \frac{1}{\sum_k x_k}
\end{align}
Now, if we notice 
\begin{equation}
    \frac{1}{x_i} + \sum_{k \in \{1, \ldots, K\} \setminus \{i\}} \frac{1}{x_k} = \sum_{k \in \{1, \ldots, K\}} \frac{1}{x_k}
\end{equation}
then we can see
\begin{align}
    \sum_k\frac{1}{x_k} \ge \frac{1}{\sum_k x_k} 
    \quad 
    \Rightarrow
    \quad
    \sum_k{x_k}  \ge \frac{1}{\sum_k\frac{1}{x_k}}\\
    \Rightarrow \sum_k IOU_k  \ge \frac{1}{\sum_k UOI_k}
\end{align}

Hence, the lower-bound of IOU has been shown as a decreasing function of UOI.
If $UOI_i=1 \,\,\, \forall i$, clearly $\sum_{k=1}^K IOU_k=K$ and $IOU=1$. This implies that UOI acts as
a good surrogate objective for optimizing IOU.
\end{proof}

\subsection{Semantic Segmentation Proposals}

Our second module is a pipeline which uses a graphical model to generate semantic segmentation proposals. 
We directly use the O$_{2}$P+\divmbest approach of \cite{yadollahpour2013rerank}.
They use the O$_{2}$P model~\cite{carreira_eccv12} which 
generates approximately 150 CPMC segments \cite{carreira_cvpr10} for each image, 
then scores them using Support Vector Regressors trained over 
second-order pooled features \cite{carreira_eccv12}. These segmentations are 
greedily pasted to form a semantic segmentation. Finally, \divmbest is used to generated 
multiple diverse semantic segmentation proposals. 
\cite{yadollahpour2013rerank} showed that one of these proposals tends to be significantly 
more accurate that the 1-best semantic segmentation. Let the \oracle segmentation be the 
most accurate segmentation in the set. 
The \oracle accuracy at just 10 proposals is $15$\pp higher than the 1-best segmentation.



\section{Post-Processing: Coarse to Full}

To evaluate segmentations from \segnet we need to post-process these coarse
segmentations to produce ``full-sized'' semantic segmentation. 
We'll refer to the up-sampled segmentation as $\hat{P}_{jk}$ where each index $j$ corresponds to a
pixel in the original image and $k$ indexes classes.
Let $\hat{p}_{ik}$ denote the probability of class $k$ predicted at coarse pixel $i$ (which corresponds to 
a patch in the full resolution image) 
by the last layer of a \segnet.

We propose four post-processing strategies that are arranged by increasing sophistication;
the first two methods are simple heuristics while the last two methods try to pick good proposals.

\textbf{(Naive)}
A simple way to do this, which we'll call 
naive upsampling just copies the argmax of a coarse pixel into all
pixels in the patch it came from:
\begin{equation} \label{eq:dump_ups}
    \hat{P}_{jk} = \argmax_{k} \hat{p}_{ik}
\end{equation}

\textbf{(Superpixel)}
The next step is to try and respect object boundaries using small superpixels (using
SLIC \cite{achanta2012slic}), which are labeled by coarse segmentations.
For each superpixel, we aggregate distributions over categories from 
patches overlapping with this superpixel. Each distribution is weighted by the percentage of 
pixels in the superpixel that are also in the patch.
This gives a distribution for superpixels. We take the $\argmax$ for each patch distribution.
Neither this smart upsampling, nor the previous \naive upsampling
are competitive.

\textbf{(\segnet)}
Next, we use \segnet outputs to pick proposals.
We down-sample each \divmbest segmentation to
$13 \times 13$ soft segmentations, similar to how ground truth was
downsampled for training.
Call $\hat{q}_{ik}^m$ the probability of the $m^{th}$ DivMBest downsampled
segmentation at patch $i$.
We score the consistency of $\hat{p}$ and $\hat{q}^m$ with the
symmetric-KL augmented by a background penalty term:
\begin{equation}
\small
S(m) = \sum_{i} \left[ D_{KL}(\hat{p}_i || \hat{q}_i^m) + D_{KL}(\hat{q}_i^m || \hat{p}_i) + 0.02 \hat{q}_{i,0}^m \right].
\end{equation}
where $D_{KL}$ is the Kullback-Leibler divergence and $0.02\hat{q}_{i, 0}^m$ is a
regularizer that penalizes background prediction.
The background penalty comes from observing that background (class 0) is frequently
overpredicted; 
adding this term consistently improved validation performance.


\textbf{(\segnet+SVM)}
The is our final approach, which works best, and uses \segnet segmentations 
as a feature, and training a \reranker to pick the best proposal from \divmbest. 
Specifically, similar to \cite{yadollahpour2013rerank}, we train a 
ranking Support Vector Machine to choose the best proposal 
according to a variety of features which describe 
proposals. We use both sophisticated hand-engineered features taken from \cite{yadollahpour2013rerank}, 
and simple features based on \segnet outputs. 
This is similar to R-CNN~\cite{girshick2013convnet} and SDS~\cite{hariharan2014sds}, which each use an SVM 
trained on CNN features to evaluate proposals.

\textbf{Segmentation features:}
\begin{enumerate}
\item[]
\textbf{(\segnet)}
As in the previous section, we calculate KL
divergence between proposals and CNN segmentations, 
and consider each direction the divergence can be computed separately
(\textbf{2} dimensions).

We also considered expected intersection, expected union,
expected intersection over expected union, and expected union over expected intersection.
Each of these statistics is computed for all PASCAL classes plus background 
(\textbf{84} dimensions). 

\item[]
\textbf{(CNN Classification)}
We use class-wise scores from an SVM
trained on DeCAF features \cite{donahue2013decaf}
for PASCAL classification (\textbf{20} dimensions).
In addition, we extend image classification to predict not just the existene of
objects, but also whether a category in an object is greather than a certain size or not.
To train an SVM for class $C$ and threshold $t \in [0, 1]$
we set the ground truth label for class $C$ to 0 if the percent of $C$
pixels in an image is below $t$.
The thresholds are chosen per-class by sorting images by percentage of $C$ pixels
then using the $C$-percent of the image at the 20th, 40th, 60th, and 80th percentile.
(\textbf{80} dimensions).

\item[]
\textbf{(DivMBest+ReRank)}
Finally, we use use all features used by \cite{yadollahpour2013rerank} (\textbf{1966} dimensions).

\end{enumerate}

%% file: figs/fig4.tex
\begin{figure}[t]

\includegraphics[width=\linewidth]{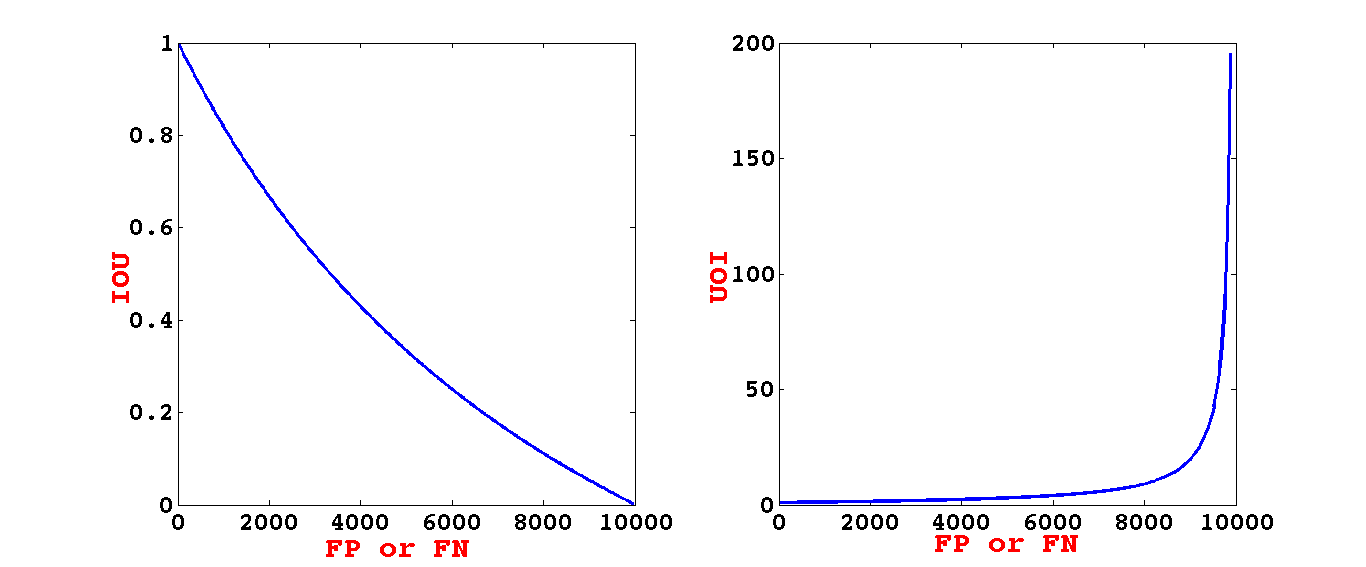}
\vspace{-1em}
    \caption{Variation of IOU and UOI losses with changes in the number of false positives and false negatives. Plots produced by flipping random pixels one by one in an original label image and computing the losses. ($FP=\sum_k FP_k = \sum_k FN_k = FN$)}
\label{fig:sim}
\vspace{-1em}
\end{figure}

%% file: results.tex
\section{Experiments and Results}

\input{figs/fig5/fig5}
\input{figs/fig6/fig6}
\input{figs/fig8/fig8}
\input{figs/table}
\input{figs/table_val}


\textbf{Setup.} 
We report our results on the PASCAL VOC 2012 segmentation dataset. 
We used the \trainval data provided by the challenge, and the additional annotations 
collected by Berkeley~\cite{bharathICCV2011}. 
We trained \segnet in a cross-val manner -- we split the entire dataset into 10 folds, 
trained \segnet on 9 folds and computed \segnet outputs in the 10th fold. 
Finally, we trained the SSVM \reranker on all training data other than \val. 
All results and analyses reported in this paper are on \val. We picked our best performing 
approach and uploaded to the PASCAL evaluation server to report results on \test.

Table \ref{tab:valresults} shows the results of all approaches on PASCAL 2012 \val set. 
\Naive upsampling performs worst at $31.3\%$. 
Superpixel upsampling gives a
small improvement at $31.9\%$. 
Neither of these are competitive, which we suspect
is due of the coarseness of the segmentations. 
It's possible that more sophisticated up-sampling strategies from  
\cite{ciresan2012deep, farabet2013pami, pinheiro2014rcnn}
would result in more competitive segmentations directly from the CNN. 
On the other hand, even our simple \reranking of DivMBest proposals is competitive, 
at $48.6\%$.  
Our final method of \segnet features with the SVM \reranker yielded best results at 
$53.1\%$. We uploaded our best performing method on the PASCAL evaluation server
and table \ref{tab:results} shows the results

Contrary to most recent CNN results, our setting allows \segnet to 
perform well with relatively little data. 
The PASCAL segmentation dataset \cite{pascal-voc-2012} 
augmented with extra annotations from \cite{bharathICCV2011} 
only has about 12000 images. 

We think a variety of decisions combined to allow competitive performance 
with such little data. 
Foremost is our ability to initialize weights for the first few layers from AlexNet, 
which was trained with a larger dataset (ImageNet).
By keeping those weights fixed, we constrained learning to the small set of
parameters contained in deeper \conv layers.
The lack of fully connected layers also helps keep the parameter count low.
This gives \segnet much less opportunity to overfit to our smaller dataset. 
Furthermore, forcing the final segmentation to be a choice from proposals 
(1) constrained the model even more (pick 1 of 30) and (2) allowed us to 
incorporate a variety of information from other methods to compensate for things 
\segnet could not learn. 
In essence, we constructed a deep model without learning a deep model. 

In figure \figref{fig:uoivsce} we show performance of \segnet \reranking
to compare losses. After training a net for 4000 iterations with cross-entropy
we continue training from that net using 3 losses. Optimizing UOI clearly
outperforms cross-entropy. Because the losses might be complementary, we
also optimize a linear comination of the two ($0.7UOI + 0.3CE$, found with grid search).
This further improves performance by a bit.

\input{figs/fig7}

\paragraph{Ablation Studies.}
We tried to tease apart the influence of different components in our pipeline. 
First, if we train an SVM \reranker with \segnet features alone, it 
performs about the same as simple 
KL divergence based ranking ($47.4\%$). 
Adding (\textbf{CNN classification}) features and (\textbf{DivMBest+ReRank}) 
features from \cite{yadollahpour2013rerank} increases this performance 
by about $3.5\%$ and $4.0\%$ respectively. 
Using both yields an extra percent of performance. 

To get a better idea of which features are considered important by SVM re-ranker,  
we considered various subsets of the features
Recall that the three types of features are (1) \segnet features, 
(2) Classification features, and 
(3) DivMBest+ReRank features from \cite{yadollahpour2013rerank}. 
For reference, using all three resulted in $53\%$ on \val.
If we only use (1) then we get $49.0\%$, only (2) gives $47.8\%$, and 
only (3) gives 
$48.1\%$. 
Thus we see that the learned \segnet features 
outperform non-segmentation CNN features (classification) 
and 1000s of dimensions of hand crafted features 
from \cite{yadollahpour2013rerank}. 
However, doing both works best. 
Using all features but \segnet features, \ie, (2)+(3) 
gives $50.5\%$, which shows that \segnet features are 
important, even in the presence of other CNN-based features. 
Using just DivMBest+ReRank features and \segnet features, \ie (1)+(3) 
performs at $52.0\%$, so \segnets again appear to be more important than simple CNN classification
features.


Some qualitative results were also interesting. In \figref{fig:bird}
we note that (probabilistic) softness of our segmentations helps alleviate 
some problems with coarseness, but we point out how such problems still 
manifest in \figref{fig:coarse} and \figref{fig:tradeoff}.

%% file: figs/fig5/fig5.tex
\begin{figure}[t]
    \begin{minipage}[c]{\columnwidth}
    \begin{subfigure}[c]{\textwidth}
        \begin{center}
        \begin{tabular}{cc}
        \begin{subfigure}{0.3\textwidth}
        \includegraphics[width=\textwidth]{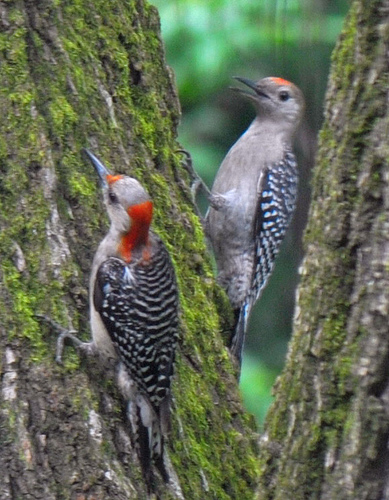}
        \end{subfigure} &
        \begin{subfigure}{0.3\textwidth}
        \begin{overpic}[width=\textwidth]{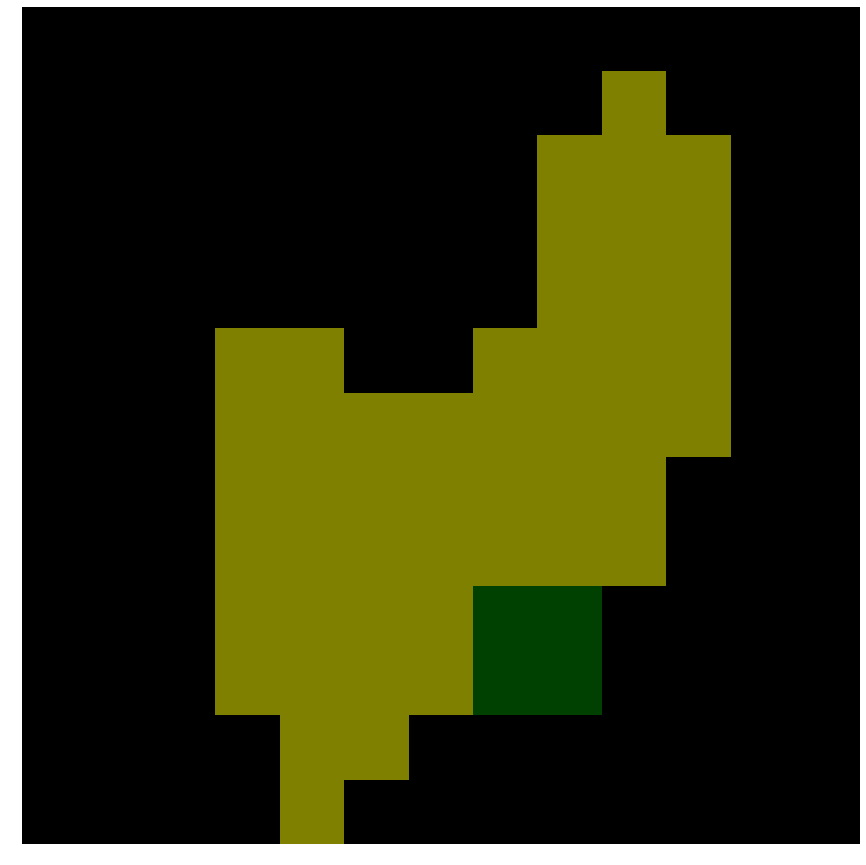}
            \put(47, 38){\color{white}\scriptsize bird}
            \put(60, 23){\color{white}\scriptsize plant}
        \end{overpic}
        \caption*{\segnet argmax}
        \end{subfigure} \\
        \begin{subfigure}{0.3\textwidth}
        \begin{overpic}[width=\textwidth]{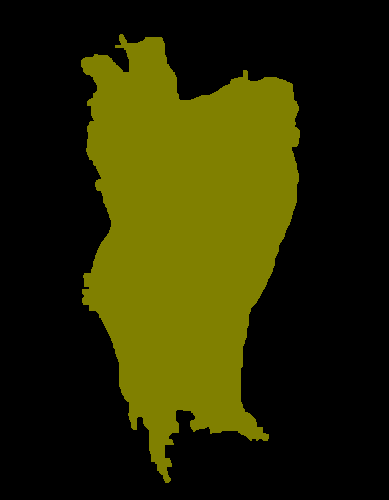}
            \put(27, 38){\color{white}\scriptsize bird}
        \end{overpic}
        \caption*{chosen}
        \end{subfigure} &
        \begin{subfigure}{0.3\textwidth}
        \begin{overpic}[width=\textwidth]{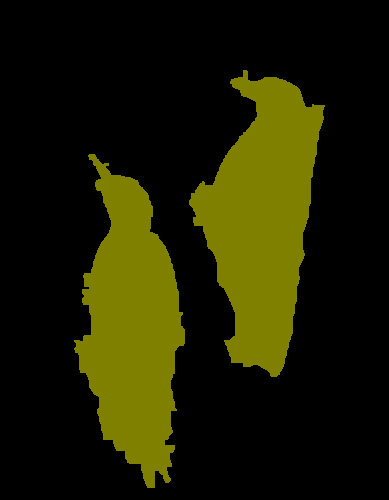}
            \put(47, 38){\color{white}\scriptsize bird}
            \put(20, 18){\color{white}\scriptsize bird}
        \end{overpic}
        \caption*{better}
        \end{subfigure} \\
        \end{tabular}
        \end{center}
    \end{subfigure}
    \end{minipage}
    \caption{If the coarse argmax were upsampled \naively then it would
    predict too much bird, so it picks a proposal
    that over-labels bird. However, if you image the argmax without a band
    of pixels around its border then it becomes too sparse,
    hence it's very hard to segment close instances like these birds
    using coarse predictions.}
    \label{fig:tradeoff}
\end{figure}

%% file: figs/fig6/fig6.tex
\begin{figure}[t]
    \begin{subfigure}[c]{\columnwidth}
        \begin{subfigure}{0.32\textwidth}
        \includegraphics[width=\textwidth]{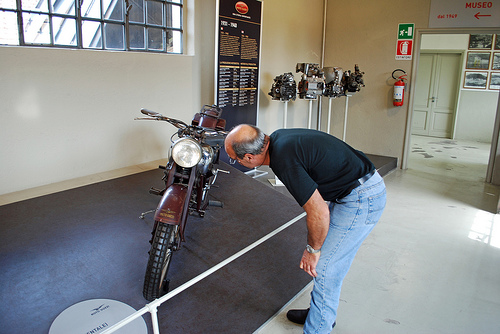}
        \caption*{Input}
        \end{subfigure}
        \begin{subfigure}{0.32\textwidth}
        \begin{overpic}[width=\textwidth]{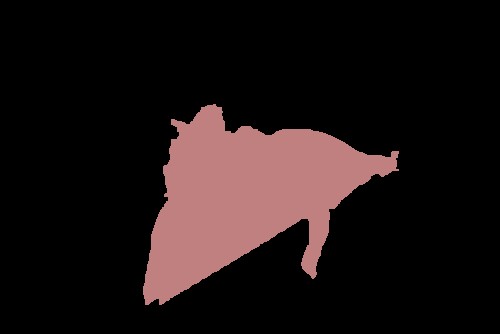}
            \put(37, 25){\color{white}\scriptsize person}
        \end{overpic}
        \caption*{\segnet}
        \end{subfigure}
        \begin{subfigure}{0.32\textwidth}
        \begin{overpic}[width=\textwidth]{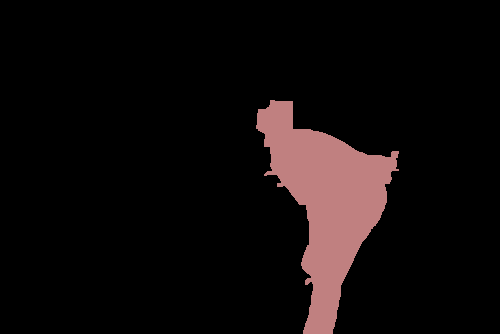}
            \put(47, 18){\color{white}\scriptsize person}
        \end{overpic}
        \caption*{\segnet+SVM}
        \end{subfigure} \\
        \begin{subfigure}{\textwidth}
        \includegraphics[width=\textwidth]{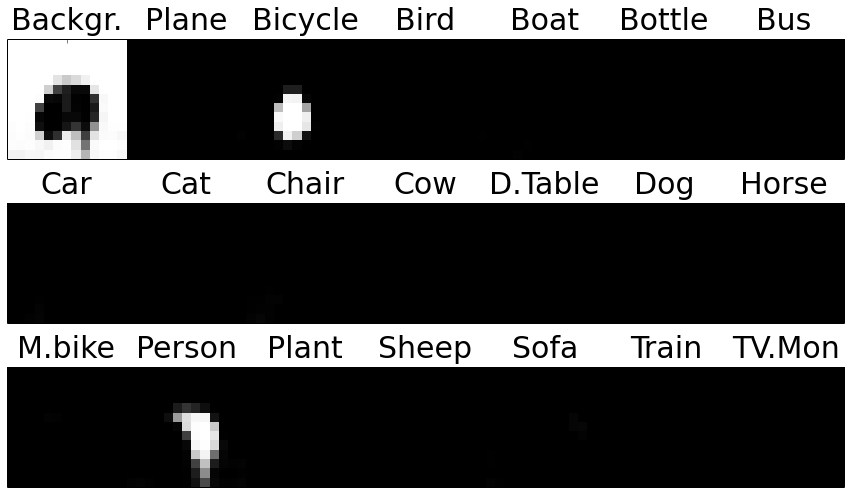}
        \caption*{\segnet prediction broken down by class}
        \end{subfigure}
    \end{subfigure}
    \caption{Here the SVM ranker both hurts and helps \segnet. The man
    bending over doesn't fit nicely into coarse patches, so \segnet can't
    tell between the non-person-like taco shape and the person hunched over
    shape. SVM ranker features make up for this since
    they're more aware of object boundaries. On the other hand, \segnet
    clearly knows that a bike is present, but it has no way of expressing
    this knowledge because the bike isn't present in any proposals.}
    \label{fig:coarse}
\end{figure}

%% file: figs/fig8/fig8.tex
\begin{figure}[t]
    \begin{minipage}[c]{\columnwidth}
    \begin{subfigure}[c]{\textwidth}
        \begin{center}
        \begin{tabular}{cc}
        \begin{subfigure}{0.3\textwidth}
        \includegraphics[width=\textwidth]{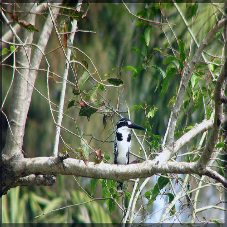}
        \end{subfigure} &
        \begin{subfigure}{0.3\textwidth}
        \begin{overpic}[width=\textwidth]{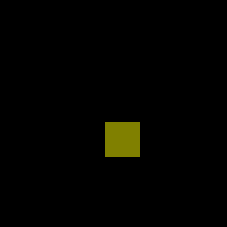}
            \put(47, 38){\color{white}\scriptsize bird}
        \end{overpic}
        \caption*{\segnet argmax}
        \end{subfigure} \\
        \begin{subfigure}{0.3\textwidth}
        \begin{overpic}[width=\textwidth]{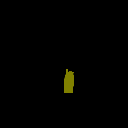}
            \put(38, 35){\color{white}\scriptsize bird}
        \end{overpic}
        \caption*{chosen}
        \end{subfigure} &
        \begin{subfigure}{0.3\textwidth}
        \begin{overpic}[width=\textwidth]{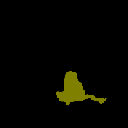}
            \put(54, 30){\color{white}\scriptsize bird}
        \end{overpic}
        \caption*{worse}
        \end{subfigure} \\
        \end{tabular}
        \end{center}
    \end{subfigure}
    \end{minipage}
    \caption{Here we can see that \segnet can reason about the size of objects.
    Even though the correct bird is small and slim, \segnet predicts a soft
    label for each of its pixels, so it can give less probability to a pixel
    if the corresponding input was small.
    Thus, it correctly chooses the small bird over the larger, inaccurate
    bird blob.}
    \label{fig:bird}
\end{figure}

%% file: figs/table.tex
\begin{table*}
\scriptsize
\begin{tabu} to \linewidth {lX[c]X[c]X[c]X[c]X[c]X[c]X[c]X[c]X[c]X[c]X[c]X[c]X[c]X[c]X[c]X[c]X[c]X[c]X[c]X[c]X[c]|X[c]}
& \begin{turn}{90}Backgr.\end{turn} &  \begin{turn}{90}Plane\end{turn} &  \begin{turn}{90}Bicycle\end{turn} &  \begin{turn}{90}Bird\end{turn} &  \begin{turn}{90}Boat\end{turn} &  \begin{turn}{90}Bottle\end{turn} &  \begin{turn}{90}Bus\end{turn} &  \begin{turn}{90}Car\end{turn} &  \begin{turn}{90}Cat\end{turn} &  \begin{turn}{90}Chair\end{turn} &  \begin{turn}{90}Cow\end{turn} &  \begin{turn}{90}D.Table\end{turn} &  \begin{turn}{90}Dog\end{turn} &  \begin{turn}{90}Horse\end{turn} &  \begin{turn}{90}M.bike\end{turn} &  \begin{turn}{90}Person\end{turn} &  \begin{turn}{90}Plant\end{turn} &  \begin{turn}{90}Sheep\end{turn} &  \begin{turn}{90}Sofa\end{turn} &  \begin{turn}{90}Train\end{turn} &  \begin{turn}{90}TV.Mon\end{turn} &  \begin{turn}{90}Average\end{turn} \\
\hline
O$_{2}$P ~\cite{carreira_eccv12} &
         84.8 & 63.7 & 23.4 & 44.9 & 40.8 & 45.1 & 58.0 & 58.8 & 57.6 & 12.1 & 43.8 & 31.0 & 44.8 & 56.2 & 56.8 & 52.3 & 37.1 & 44.0 & 29.5 & 48.6 & 42.9 & 46.5 \\
O$_{2}$P DivMBest+ReRank ~\cite{yadollahpour2013rerank} &
         85.7 & 62.7 & 25.6 & 46.9 & 43.0 & 54.8 & 58.4 & 58.6 & 55.6 & 14.6 & 47.5 & 31.2 & 44.7 & 51.0 & 60.9 & 53.5 & 36.6 & 50.9 & 30.1 & 50.2 & 46.8 & 48.1 \\
UDS ~\cite{dong2014nusseg} & 85.2 & 67.0 & 24.5 & 47.2 & \textbf{45.0} & 47.9 & 65.3 & 60.6 & \textbf{58.5} & 15.5 & 50.8 & 37.4 & 45.8 & \textbf{59.9} & 62.0 & 52.7 & 40.8 & 48.2 & \textbf{36.8} & 53.1 & 45.6 & 50.0 \\
SDS ~\cite{hariharan2014sds} & 86.7 & 63.3 & 25.7 & \textbf{63.0} & 39.8 & \textbf{59.2} & \textbf{70.9} & \textbf{61.4} & 54.9 & \textbf{16.8} & 45.0 & \textbf{48.2} & 50.5 & 51.0 & 57.7 & \textbf{63.3} & 31.8 & 58.7 & 31.2 & \textbf{55.7} & \textbf{48.5} & 51.6 \\
\segnets & \textbf{86.8} & \textbf{70.2} & \textbf{27.0} & 57.6 & 44.6 & 54.0 & 69.0 & 58.5 & 56.6 & 14.6 & \textbf{59.3} & 34.5 & \textbf{52.3} & 59.5 & \textbf{64.2} & 59.1 & \textbf{41.3} & \textbf{61.6} & 33.8 & 51.8 & 46.8 &  \textbf{52.5} \\
\hline
\end{tabu}
\caption{
PASCAL VOC 2012 segmentation \test results. ~\cite{carreira_eccv12} is the same as picking the highest scoring DivMBest solution.}
\label{tab:results}
\end{table*}

%% file: figs/table_val.tex
\begin{table*}[t]
\centering
\begin{tabular}{@{\extracolsep{\fill}}cccccc@{\extracolsep{\fill}}}
\toprule
\Naive Upsampling & Superpixel Upsampling & \segnet-CE & \segnet-UOI & \segnet-UOI-CE-combination & +SVM \reranker\\
\hline
31.3  &  31.9  &  47.4  &  48.5  &  48.7  &  53.1 \\
\hline
\end{tabular}
\caption{PASCAL VOC 2012 segmentation \val results.} 
\label{tab:valresults}
\end{table*}

%% file: figs/fig7.tex
\begin{figure}[t]

\includegraphics[width=\linewidth]{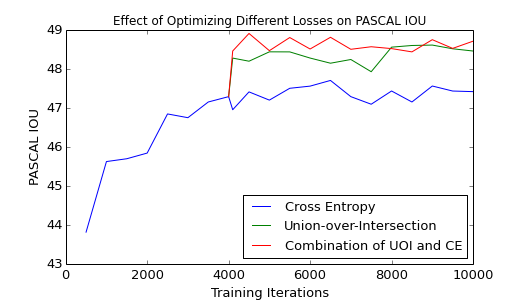}
\vspace{-1em}
    \caption{This plot shows performance of \segnet \reranking for \segnets trained
    with different loss types. Training with UOI helps pick better solutions
    and training with a linear combination of UOI and cross entropy helps
    even more.}
\label{fig:uoivsce}
\vspace{-1em}
\end{figure}

%% file: conclusion.tex
\section{Conclusion and Future Work}

To summarize, we present a two-module approach for semantic segmentation. 
Module 1 uses a graphical model to produce 
multiple semantic segmentation proposals. 
Module 2 uses \textbf{\segnet}, a novel CNN which 
is specifically trained for semantic segmentation task-loss, and is 
used to to score then \rerank these proposals, resulting in a final segmentation. 
Our approach achieves $52.5\%$ on the PASCAL 2012 segmentation challenge.

Our experiments with and without proposals 
reach findings that are consistent with those observed in previous work -- 
that methods which use proposals consistently outperform those which don't, even
among methods which use CNNs. 
In object detection non-proposals methods (sliding window) still 
are efficient enough to be viable for CNNs \cite{sermanet2013overfeat}. 
However, for segmentation, proposals are vital. 

Perhaps unsurprisingly, it is clear that more work needs to be done with CNNs.
Our results suggest that some information is not captured by our
CNN. 
Achieving peak performance in our pipeline requires
hand crafted features from \cite{yadollahpour2013rerank}, though our \segnet-based 
features are significant contributors. 

Another path of future experiments involves Microsoft's Common Objects in
Context dataset \cite{coco}, which contains an order of magnitude more
segmentation ground truth than has existed before. 
Following AlexNet, it seems reasonable to expect better performance from a larger  
net trained on a bigger dataset. Since our net predicts entire segmentations
instead of single labels it could have more to learn from this dataset than 
classification nets.

%% file: uoi_derivative/uoi_loss_gradient_innards.tex
\title{ \vspace{-9ex} Gradients for Optimizing Structured Segmentation Losses \vspace{-2ex}  %
}

\author{}
\date{}

\maketitle

\section{Introduction and Notation}

This document is meant to accompany the paper ``Combining the Best of Graphical
Models and ConvNets for Semantic Segmentation''. Herein we compute the derivatives
of Intersection-over-Union (IOU) and Union-over-Intersection (UOI) with respect
to an input feature map $\mathbf{z}$. Two forms of the IOU derivative can be
seen in \eqref{eq:IOU_expect} and \eqref{eq:IOU_prob}. Corresponding forms
of the UOI derivative are in \eqref{eq:UOI_expect} and \eqref{eq:UOI_prob}.

We'd like to use a supervised learning algorithm to train a model for labeling
image pixels using $K$ classes. Let $\bm{\hat{Y}}$ be the predicted
segmentation and $\bm{Y}$ the ground truth so that $\hat{Y}_i$ and $Y_i$ are
random variables indicating the class of pixel $i$; each takes values in
$\{1, \ldots, K\}$ and pixels labeled with indices $\{1, \ldots, N\}$. The set of pixels
can either be all pixels in an image or all pixels in a minibatch, though we take
it to be the set of all pixels in a minibatch. Thus,
$P(\hat{Y}_i = k)$ is the probability of predicting pixel $i$ is class $k$.
However, we'll introduce the following short hand to condense the notation:
\begin{equation}
    \hat{p}_{i, k} = P(\hat{Y}_i = k)
\end{equation}
and
\begin{equation}
    p_{i, k} = P(Y_i = k)
\end{equation}

Here, the model produces a score $z_{i, k}$ for each pixel $i$ and class $k$
then predicts $\bm{\hat{Y}}$ using a softmax. It assigns
\begin{equation}
    \hat{p}_{i, k}
        := \frac{\exp(z_{i, k})}{\sum_{\tilde{k}} \exp(z_{i, \tilde{k}})}
\end{equation}

First we'll build up some machinery useful for both IOU and UOI, then we'll
compute their derivatives.

\section{Intersection and Union}

The intersection function captures the notion of agreement between ground
truth and prediction. It's defined for class $k \in \{1, \ldots, K\}$ as
\begin{equation}
    I_k(\bm{\hat{Y}}, \bm{Y}) = \sum_i^N \ind{\hat{Y}_i = k \land Y_i = k}
\end{equation}
where $\ind{\cdot}$ is the indicator function.
We're interested in the expected intersection,
\begin{align}
    E\left[ I_k(\bm{\hat{Y}}, \bm{Y}) \right]
        &= E\left[ \sum_i^N \ind{\hat{Y}_i = k \land Y_i = k} \right] \\
        &= \sum_i^N E\left[ \ind{\hat{Y}_i = k \land Y_i = k} \right]
\end{align}
which can be reduced to the following, because the expectation of an indicator
function is the probability of the event inside:
\begin{align}
    E\left[ I_k(\bm{\hat{Y}}, \bm{Y}) \right]
        &= \sum_i^N \hat{p}_{i, k} p_{i, k}
\end{align}

The union function is about the total "footprint" in pixels of ground truth and
prediction; it is defined as
\begin{equation}
    U_k(\bm{\hat{Y}}, \bm{Y}) = \sum_i^N \ind{\hat{Y}_i = k \lor Y_i = k}
\end{equation}
Expected union is analogous to expected intersection:
\begin{align}
    E\left[ U_k(\bm{\hat{Y}}, \bm{Y}) \right]
        &= E\left[ \sum_i^N \ind{\hat{Y}_i = k \lor Y_i = k} \right] \\
        &= \sum_i^N E\left[ \ind{\hat{Y}_i = k \lor Y_i = k} \right] \\
        &= \sum_i^N \left( \hat{p}_{i, k} + p_{i, k} - \hat{p}_{i, k} p_{i, k} \right)
\end{align}

In fact, it can be expressed using Intersection (inclusion-exclusion principle)
\begin{align}
    E\left[ U_k(\bm{\hat{Y}}, \bm{Y}) \right]
        &= \sum_i^N \left( \hat{p}_{i, k} + p_{i, k} \right) - E\left[ I_k(\bm{\hat{Y}}, \bm{Y}) \right]
\end{align}

\section{Softmax Gradient}

A softmax function takes a bunch of scores and outputs probabilities.
In the next section we'll need the softmax's gradient at any output with
respect to any input. Normally we care about the gradient of an output with
respect its corresponding input, but this case is a bit more general.
Here $k' \in \{1, \ldots, K\}$.

\begin{align}
    \frac{\del \hat{p}_{i, k'}}{\del z_{i, k}}
        &= \frac{\del}{\del z_{i, k}} \frac{\exp(z_{i, k'})}{\sum_{\tilde{k}} \exp(z_{i, \tilde{k}})} \\
        &= \frac{\frac{\del}{\del z_{i, k}} \left( \exp(z_{i, k'}) \right) \sum_{\tilde{k}} \exp(z_{i, \tilde{k}}) -
                 \exp(z_{i, k'}) \frac{\del}{\del z_{i, k}} \left( \sum_{\tilde{k}} \exp(z_{i, \tilde{k}}) \right)}
                {\left( \sum_{\tilde{k}} \exp(z_{i, \tilde{k}}) \right)^2} \\
\end{align}

If $k = k'$ then
\begin{align}
    \frac{\del \hat{p}_{i, k'}}{\del z_{i, k}}
        &= \frac{\exp(z_{i, k}) \sum_{\tilde{k}} \exp(z_{i, \tilde{k}}) -
                 \exp(z_{i, k}) \exp(z_{i, k})}
                {\left( \sum_{\tilde{k}} \exp(z_{i, \tilde{k}}) \right)^2} \\
        &= \hat{p}_{i, k} - \hat{p}_{i, k}^2
\end{align}

When $k \ne k'$,
\begin{align}
    \frac{\del \hat{p}_{i, k'}}{\del z_{i, k}}
        &= \frac{- \exp(z_{i, k'}) \exp(z_{i, k})}
                {\left( \sum_{\tilde{k}} \exp(z_{i, \tilde{k}}) \right)^2} \\
        &= - \hat{p}_{i, k'} \hat{p}_{i, k}
\end{align}

Now the whole derivative can be written as one expression and simplified a bit.
\begin{align}
    \frac{\del \hat{p}_{i, k'}}{\del z_{i, k}}
        &= \ind{k = k'} (\hat{p}_{i, k} - \hat{p}_{i, k}^2) -
           \ind{k \ne k'} \hat{p}_{i, k'} \hat{p}_{i, k} \\
\end{align}

In the first case, substitute $k'$ for $k$ to get
\begin{align}
\ind{k = k'} (\hat{p}_{i, k} - \hat{p}_{i, k}^2) -
   \ind{k \ne k'} \hat{p}_{i, k'} \hat{p}_{i, k}
        &= \ind{k = k'} \hat{p}_{i, k'} (1 - \hat{p}_{i, k}) -
           \ind{k \ne k'} \hat{p}_{i, k'} \hat{p}_{i, k} \\
        &= \hat{p}_{i, k'} \left( \ind{k = k'} (1 - \hat{p}_{i, k}) +
                                  \ind{k \ne k'} (-\hat{p}_{i, k}) \right) \\
        &= \hat{p}_{i, k'} \left( \ind{k = k'} (1 - \hat{p}_{i, k}) +
                                  \ind{k \ne k'} (0 - \hat{p}_{i, k}) \right) \\
        &= \hat{p}_{i, k'} (\ind{k = k'} - \hat{p}_{i, k})
\end{align}

\section{Gradient of Expected Intersection and Expected Union}

Next, compute the derivatives of expected intersection and expected union.
For intersection,
\begin{align}
    \frac{\del}{\del \hat{p}_{i, k}} E[I_{k'}(\bm{\hat{Y}}, \bm{Y})]
        &= \frac{\del}{\del \hat{p}_{i, k}} \sum_j^N \hat{p}_{j, k'} p_{j, k'} \\
        &= p_{i, k'} \frac{\del \hat{p}_{i, k'}}{\del \hat{p}_{i, k}} \label{eq:I_grad}
\end{align}

In the case of union,
\begin{align}
    \frac{\del}{\del \hat{p}_{i, k}} E[U_{k'}(\bm{\hat{Y}}, \bm{Y})]
        &= \frac{\del}{\del \hat{p}_{i, k}} \sum_j^N \left( \hat{p}_{j, k'} + p_{j, k'} - \hat{p}_{j, k'} p_{j, k'} \right) \\
        &= (1 - p_{i, k'}) \frac{\del \hat{p}_{i, k'}}{\del \hat{p}_{i, k}} \label{eq:U_grad}
\end{align}

\section{IOU Loss and its Gradient}

Now, write out the IOU loss function
\begin{equation} \label{eq:exp_iou}
    \mathcal{L}_{IOU}(\bm{\hat{Y}}, \bm{Y}) = \sum_{k'}^K \frac{E[I_{k'}(\bm{\hat{Y}}, \bm{Y})]}{E[U_{k'}(\bm{\hat{Y}}, \bm{Y})]}
\end{equation}

and compute the gradient of IOU 
\begin{align}
    \frac{\del \mathcal{L}_{IOU}(\bm{\hat{Y}}, \bm{Y})}{\del z_{i, k}}
        &= \left( \sum_{k'}^K \frac{\del}{\del z_{i, k}} \frac{E[I_{k'}(\bm{\hat{Y}}, \bm{Y})]}{E[U_{k'}(\bm{\hat{Y}}, \bm{Y})]} \right) \\
        &= \sum_{k'}^K \left( \frac{\del}{\del \hat{p}_{i, k}} \frac{E[I_{k'}(\bm{\hat{Y}}, \bm{Y})]}{E[U_{k'}(\bm{\hat{Y}}, \bm{Y})]}
            \frac{\del \hat{p}_{i, k}}{\del z_{i, k}} \right) \label{eq:IOU_chain}
\end{align}

First we'll focus on 
\begin{align}
\frac{\del}{\del \hat{p}_{i, k}} \frac{E[I_{k'}(\bm{\hat{Y}}, \bm{Y})]}{E[U_{k'}(\bm{\hat{Y}}, \bm{Y})]}
\end{align}

By substituting these into the following we get
\begin{align}
    \frac{\del}{\del \hat{p}_{i, k}} \frac{E[I_{k'}(\bm{\hat{Y}}, \bm{Y})]}{E[U_{k'}(\bm{\hat{Y}}, \bm{Y})]}
        &= \frac{
             E[U_{k'}(\bm{\hat{Y}}, \bm{Y})] \frac{\del}{\del \hat{p}_{i, k}} E[I_{k'}(\bm{\hat{Y}}, \bm{Y})]
             - E[I_{k'}(\bm{\hat{Y}}, \bm{Y})] \frac{\del}{\del \hat{p}_{i, k}} E[U_{k'}(\bm{\hat{Y}}, \bm{Y})]
           }
           {E[U_{k'}(\bm{\hat{Y}}, \bm{Y})]^2} \\
        &= \frac{
             E[U_{k'}(\bm{\hat{Y}}, \bm{Y})] p_{i, k'} \frac{\del \hat{p}_{i, k'}}{\del \hat{p}_{i, k}} - E[I_{k'}(\bm{\hat{Y}}, \bm{Y})] (1 - p_{i, k'}) \frac{\del \hat{p}_{i, k'}}{\del \hat{p}_{i, k}}
           }
           {E[U_{k'}(\bm{\hat{Y}}, \bm{Y})]^2} \\
        &= \frac{
             E[U_{k'}(\bm{\hat{Y}}, \bm{Y})] p_{i, k'} - E[I_{k'}(\bm{\hat{Y}}, \bm{Y})] (1 - p_{i, k'})
           }
           {E[U_{k'}(\bm{\hat{Y}}, \bm{Y})]^2} \frac{\del \hat{p}_{i, k'}}{\del \hat{p}_{i, k}}
           \label{eq:IU_version}
\end{align}

Finally, substituting \eqref{eq:IU_version} into \eqref{eq:IOU_chain} gives
\begin{align}
    \frac{\del \mathcal{L}_{IOU}(\bm{\hat{Y}}, \bm{Y})}{\del z_{i, k}}
    &= \sum_{k'}^K \left( \frac{\del}{\del \hat{p}_{i, k}} \frac{E[I_{k'}(\bm{\hat{Y}}, \bm{Y})]}{E[U_{k'}(\bm{\hat{Y}}, \bm{Y})]}
        \frac{\del \hat{p}_{i, k}}{\del z_{i, k}} \right) \\
    &= \sum_{k'}^K \left( \frac{E[U_{k'}(\bm{\hat{Y}}, \bm{Y})] p_{i, k'}
             - E[I_{k'}(\bm{\hat{Y}}, \bm{Y})] (1 - p_{i, k'})}
            {E[U_{k'}(\bm{\hat{Y}}, \bm{Y})]^2} \frac{\del \hat{p}_{i, k'}}{\del \hat{p}_{i, k}}
        \frac{\del \hat{p}_{i, k}}{\del z_{i, k}} \right) \\
    &= \sum_{k'}^K \left( \frac{E[U_{k'}(\bm{\hat{Y}}, \bm{Y})] p_{i, k'}
             - E[I_{k'}(\bm{\hat{Y}}, \bm{Y})] (1 - p_{i, k'})}
            {E[U_{k'}(\bm{\hat{Y}}, \bm{Y})]^2} \frac{\del \hat{p}_{i, k'}}{\del \hat{z}_{i, k}} \right) \\
    &= \sum_{k'}^K \left( \frac{E[U_{k'}(\bm{\hat{Y}}, \bm{Y})] p_{i, k'}
             - E[I_{k'}(\bm{\hat{Y}}, \bm{Y})] (1 - p_{i, k'})}
            {E[U_{k'}(\bm{\hat{Y}}, \bm{Y})]^2}
       \hat{p}_{i, k'} (\ind{k = k'} - \hat{p}_{i, k}) \right) \label{eq:IOU_expect} \\
    &= \sum_{k'}^K \left( \frac{p_{i, k'} \sum_j^N \left( \hat{p}_{j, k'} + p_{j, k'} - \hat{p}_{j, k'} p_{j, k'} \right)
             - (1 - p_{i, k'}) \sum_j^N \left( \hat{p}_{j, k'} p_{j, k'} \right) }
            {\left( \sum_j^N \left( \hat{p}_{j, k'} + p_{j, k'} - \hat{p}_{j, k'} p_{j, k'} \right) \right)^2}
       \hat{p}_{i, k'} (\ind{k = k'} - \hat{p}_{i, k}) \right) \\
    &= \sum_{k'}^K \left( \frac{p_{i, k'} \sum_j^N \left( \hat{p}_{j, k'} + p_{j, k'}\right)
             - \sum_j^N \left( \hat{p}_{j, k'} p_{j, k'} \right) }
            {\left( \sum_j^N \left( \hat{p}_{j, k'} + p_{j, k'} - \hat{p}_{j, k'} p_{j, k'} \right) \right)^2}
       \hat{p}_{i, k'} (\ind{k = k'} - \hat{p}_{i, k}) \right) \label{eq:IOU_prob}
\end{align}

\section{Union over Intersection}

The gradient of Union over Intersection can be computed in a similar fashion.
UOI loss is defined as
\begin{equation} \label{eq:exp_uoi}
    \mathcal{L}_{UOI}(\bm{\hat{Y}}, \bm{Y}) = \sum_{k'}^K \frac{E[U_{k'}(\bm{\hat{Y}}, \bm{Y})]}{E[I_{k'}(\bm{\hat{Y}}, \bm{Y})]}
\end{equation}

Given \eqref{eq:I_grad} and \eqref{eq:U_grad}, we can compute the derivative of UOI:
\begin{align}
    \frac{\del \mathcal{L}_{UOI}(\bm{\hat{Y}}, \bm{Y})}{\del z_{i, k}}
    &= \sum_{k'}^K \left( \frac{\del}{\del \hat{p}_{i, k}} \frac{E[U_{k'}(\bm{\hat{Y}}, \bm{Y})]}{E[I_{k'}(\bm{\hat{Y}}, \bm{Y})]}
        \frac{\del \hat{p}_{i, k}}{\del z_{i, k}} \right) \\
    &= \sum_{k'}^K \left( \frac{E[I_{k'}(\bm{\hat{Y}}, \bm{Y})] \frac{\del}{\del \hat{p}_{i, k}} E[U_{k'}(\bm{\hat{Y}}, \bm{Y})]
             - E[U_{k'}(\bm{\hat{Y}}, \bm{Y})] \frac{\del}{\del \hat{p}_{i, k}} E[I_{k'}(\bm{\hat{Y}}, \bm{Y})]}
            {E[I_{k'}(\bm{\hat{Y}}, \bm{Y})]^2}
        \frac{\del \hat{p}_{i, k}}{\del z_{i, k}} \right) \\
    &= \sum_{k'}^K \left( \frac{E[I_{k'}(\bm{\hat{Y}}, \bm{Y})] (1 - p_{i, k'})
             - E[U_{k'}(\bm{\hat{Y}}, \bm{Y})] p_{i, k'}}
            {E[I_{k'}(\bm{\hat{Y}}, \bm{Y})]^2} \frac{\del \hat{p}_{i, k'}}{\del \hat{p}_{i, k}}
        \frac{\del \hat{p}_{i, k}}{\del z_{i, k}} \right) \\
    &= \sum_{k'}^K \left( \frac{E[I_{k'}(\bm{\hat{Y}}, \bm{Y})] (1 - p_{i, k'})
             - E[U_{k'}(\bm{\hat{Y}}, \bm{Y})] p_{i, k'}}
            {E[I_{k'}(\bm{\hat{Y}}, \bm{Y})]^2} \frac{\del \hat{p}_{i, k'}}{\del \hat{z}_{i, k}} \right) \\
    &= \sum_{k'}^K \left( \frac{E[I_{k'}(\bm{\hat{Y}}, \bm{Y})] (1 - p_{i, k'})
             - E[U_{k'}(\bm{\hat{Y}}, \bm{Y})] p_{i, k'}}
            {E[I_{k'}(\bm{\hat{Y}}, \bm{Y})]^2}
       \hat{p}_{i, k'} (\ind{k = k'} - \hat{p}_{i, k}) \right) \label{eq:UOI_expect} \\
    &= \sum_{k'}^K \left( \frac{(1 - p_{i, k'}) \sum_j^N \left( \hat{p}_{j, k'} p_{j, k'} \right)
             - p_{i, k'} \sum_j^N \left( \hat{p}_{j, k'} + p_{j, k'} - \hat{p}_{j, k'} p_{j, k'} \right)}
            {\left( \sum_j^N \left( \hat{p}_{j, k'} p_{j, k'} \right) \right)^2}
       \hat{p}_{i, k'} (\ind{k = k'} - \hat{p}_{i, k}) \right) \\
    &= \sum_{k'}^K \left( \frac{\sum_j^N \left( \hat{p}_{j, k'} p_{j, k'} \right)
             - p_{i, k'} \sum_j^N \left( \hat{p}_{j, k'} + p_{j, k'}\right)}
            {\left( \sum_j^N \left( \hat{p}_{j, k'} p_{j, k'} \right) \right)^2}
       \hat{p}_{i, k'} (\ind{k = k'} - \hat{p}_{i, k}) \right) \label{eq:UOI_prob}
\end{align}